\documentclass[11pt]{article}
\usepackage[margin=1in]{geometry}
\usepackage{amsmath,amssymb}
\usepackage{booktabs}
\usepackage{graphicx}
\usepackage{algorithm}
\usepackage{algpseudocode}
\usepackage{hyperref}

\title{Multi-Group Pipe Routing under Permanent Geometric Occupancy:\\
Problem, Benchmark, and Classical Baselines}
\author{Deng}
\date{}

\begin{document}
\maketitle


\begin{abstract}
Additive manufacturing (AM) enables compact hydraulic components whose internal
fluid channels can follow free-form 3D paths rather than conventionally drilled
holes. A representative case is multi-group channel layout in rotary
direct-drive servo valves: once a channel is placed it permanently occupies
volume, so later channels must clear earlier geometry---unlike classical
multi-agent pathfinding (MAPF), where agents free space after moving.
We study this setting as \emph{multi-group pipe routing under permanent
geometric occupancy}. Our contributions are a problem formalization with
geometric dual-witness conflicts, a constructive 3D benchmark (two corridor
generators $\times$ two obstacle painters, controlled difficulty, feasibility
witnesses), and baseline results for CBS, PBS, and priority planning adapted to
this coupling. We evaluate by success rate under a fixed time budget.
\textsc{PlaneSlice} hard instances clearly rank
$\mathrm{CBS}\gg\mathrm{PBS}\gg\mathrm{PP}$, while easier cells validate the
pipeline. The goal is a reproducible problem definition, suite, and classical
baselines---not a new optimal MAPF algorithm.
\end{abstract}


\section{Introduction and Motivation}
\label{sec:intro}

Conventional hydraulic valves often realize internal flow paths by
intersecting drilled holes. Additive manufacturing removes much of that
machining constraint: channels can be shorter, smoother, and spatially
interleaved inside a dense envelope, which is attractive for reducing
pressure loss and package size.

A motivating device is the \emph{high-performance rotary direct-drive servo
valve}. Structurally it combines:
\begin{itemize}
  \item a \textbf{center axis core} driven by a motor (rotary spool / sleeve
        interface);
  \item several \textbf{groups of external terminals} (ports) on the valve
        body;
  \item \textbf{non-conflicting fluid pipes} that connect terminals of each
        group toward the center axis (and related internal openings).
\end{itemize}
At different rotation angles, different groups are brought into hydraulic
connection through the rotary core. Therefore pipes belonging to
\emph{different} groups must remain geometrically separated for all design
poses of interest, while pipes \emph{within} a group may share trunks or
branch (tree-like manifolds)%
\IfFileExists{figures/rotary_valve_pipes.png}{, as illustrated in
Fig.~\ref{fig:rotary-valve}}{}.

\IfFileExists{figures/rotary_valve_pipes.png}{%
\begin{figure}[t]
  \centering
  \includegraphics[width=0.72\linewidth]{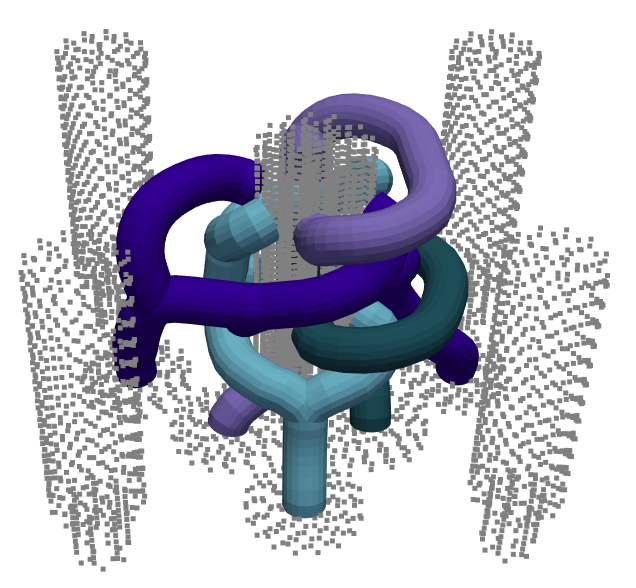}
  \caption{Illustrative multi-group fluid channels around a center axis and
  structural pillars (visualisation of an AM-oriented valve-style layout).
  Coloured tubes are pipes of different groups; grey volumes mark occupied /
  forbidden structure. Channels are short and smoothly curved relative to
  drilled manifolds, but must not collide.}
  \label{fig:rotary-valve}
\end{figure}
}{%
}

AM makes such layouts manufacturable, but \emph{designing} the channels is
still a combinatorial 3D routing problem: many terminals, cluttered obstacles,
and hard clearance between groups. Manual CAD routing does not scale; we seek
automatic planners that return conflict-free multi-group layouts inside a
bounded design domain.

\paragraph{Contributions.}
\begin{enumerate}
  \item \textbf{Problem.} We formalize multi-group pipe routing under permanent
        geometric occupancy, with dual-witness clearance conflicts and
        topology-driven multi-terminal groups (Sec.~\ref{sec:problem}).
  \item \textbf{Benchmark.} We construct a synthetic 3D suite with known-feasible
        witnesses, two corridor generators, two obstacle painters, and
        easy/middle/hard schedules (Sec.~\ref{sec:benchmark}).
  \item \textbf{Baselines.} We adapt CBS, PBS, and priority planning to this
        stack and report fixed-budget success rates on $36$ batches of
        $N{=}100$ instances each (Secs.~\ref{sec:method}--\ref{sec:experiments}).
\end{enumerate}
We do not claim a new optimal high-level algorithm or industrial CAD parity.

\section{From Valve Design to a Computational Problem}
\label{sec:problem}

We abstract the valve setting as \emph{multi-group pipe routing under permanent
geometric occupancy}. The design domain is a bounded 3D region with static
obstacles (housing, fasteners, sealed volumes, axis envelope, \ldots).
Terminals are partitioned into groups that may share a manifold toward the
axis; each routed pipe permanently occupies a tubular neighbourhood of radius
$r$ (there is no ``pass and release'' as in robot MAPF); cross-group geometric
interference is forbidden, while within-group sharing / branching is allowed
via a topology plan. This combination is not captured by classical MAPF
suites, so we formalize the problem below and construct a synthetic benchmark
(Sec.~\ref{sec:benchmark}).

\paragraph{Geometry and occupancy.}
The domain is discretised as a regular 3D grid of resolution $\Delta$
(we use $\Delta{=}1$ without loss of generality by scaling).
A pipe of radius $r$ occupies a \emph{capsule} neighbourhood along its
centreline (sphere swept along each segment). Static obstacles are modelled as
spheres, capsules, and/or axis-aligned boxes.
A query point (or cell centre) $p$ with pipe radius $r$ is \textbf{free} iff
its clearance to all static obstacles and to already committed pipes of
\emph{other} groups is at least $r$. Equivalently, the open ball $B(p,r)$ must
not intersect forbidden geometry.

\paragraph{Groups, terminals, and topology.}
An instance specifies $m$ groups $G_1,\ldots,G_m$. Group $G_i$ has terminals
$T_i=\{t_{i,1},\ldots,t_{i,n_i}\}$ on the domain boundary or on prescribed
interfaces such as the axis sleeve (ports in the valve metaphor; center-axis
openings may be additional terminals or topology goals).
Before geometric search, a topology builder maps $T_i$ to tasks (edges)
$E_i$---typically a spanning tree---and an optional intra-group growth order.
Solving all tasks of $G_i$ yields a connected tree-shaped network.
Cross-group tasks interact only through occupancy.

\paragraph{Conflicts.}
Unlike classical MAPF vertex/edge conflicts at a shared time, two pipes of
different groups conflict when their tubular neighbourhoods violate clearance.
A detected conflict stores witness points $(p_a,r_a)$ and $(p_b,r_b)$
(generally $p_a\neq p_b$) and a deficit measuring penetration.
Same-group contacts are not conflicts under the default sharing policy.

\paragraph{Feasible solution and objectives.}
A feasible solution is a set of paths for all tasks of all groups such that
every path is free w.r.t.\ static obstacles and all other-group committed
geometry.
The primary experimental objective in this paper is \emph{feasibility under a
wall-clock budget} (success rate). Secondary signals include total path cost
relative to a constructive witness.
Manufacturing-oriented criteria (bend count, supportability, pressure-loss
proxies) matter for valves but are left to future work; we focus on the
geometric multi-group core first.

\paragraph{Relation to MAPF.}
\begin{table}[htbp]
  \centering
  \caption{Classical MAPF vs.\ multi-group pipe routing under permanent
  occupancy.}
  \label{tab:vs-mapf}
  \begin{tabular}{@{}llp{0.36\linewidth}@{}}
    \toprule
    Aspect & Classical MAPF & This problem \\
    \midrule
    Occupancy & Temporary (agents move) & Permanent (pipes stay) \\
    Conflict & Same cell / edge @ time $t$ & Geometric clearance (dual witnesses) \\
    Agent & Single $s$--$t$ path & Multi-terminal \textbf{group} (tree) \\
    Time & Explicit time dimension & Design-time layout (no schedule) \\
    \bottomrule
  \end{tabular}
\end{table}
As shown in Table~\ref{tab:vs-mapf}, the settings differ in occupancy, conflict
semantics, and agent structure.
High-level MAPF controllers (CBS, PBS, priority planning) remain useful as
search over leaves (constraint sets or orders), provided low-level search and
conflict detection use permanent geometric occupancy.
We do not claim a drop-in replacement for spatiotemporal MAPF benchmarks.


\section{Baseline Planner Stack}
\label{sec:method}

We adapt CBS, PBS, and priority planning (PP) to permanent geometric
occupancy and multi-terminal groups.
The shared stack comprises four layers:

\begin{description}
  \item[Topology.]
  Terminals $\{T_i\}$ map to tasks (edges) and an optional intra-group growth
  order.
  Reported runs use a Euclidean \emph{pure MST} per group: same-group routed
  segments become connectable goals, forming a tree-shaped manifold.
  Cross-group interaction is only through occupancy.

  \item[Occupancy.]
  Static geometry uses a clearance grid; committed other-group pipes and CBS
  negative constraints use an \emph{obstacle chain} of spheres and capsules.
  A cell centre $c$ is free for radius $r$ iff clearance to every forbidden
  volume is at least $r$.

  \item[Low-level search.]
  Each task is solved by grid A$^\star$ (26-connected, Manhattan) on the
  composite grid, yielding a free cell polyline.
  Under PP the path is committed as hard occupancy for later groups; under
  CBS/PBS it is held in the current leaf solution.

  \item[Conflict detection.]
  After all groups are routed in a leaf, we detect cross-group geometric hits
  as dual witnesses $(p_a,r_a)$, $(p_b,r_b)$ with a deficit (same-group contacts
  ignored).
  CBS/PBS rank conflicts with a \emph{near-terminal} heuristic when choosing
  which hit to resolve.
\end{description}

\subsection{High-level controllers}
\label{sec:high-level}

\begin{description}
  \item[CBS.]
  Constraint-tree leaves store per-group paths plus negative witness spheres on
  the chain.
  A selected conflict adds opposing constraints; only the affected group is
  replanned.

  \item[PBS.]
  Leaves store a priority DAG over groups.
  A conflict adds a priority edge; the lower-priority group replans with
  higher-priority paths as hard obstacles
  (reported: depth-first expansion).

  \item[PP.]
  A total order over groups is searched: prior groups are hard obstacles for
  later ones; failure triggers sequence neighbourhood repair.
  There is no dual-witness constraint branching---only order search.
\end{description}


\section{Benchmark Construction}
\label{sec:benchmark}

Classical MAPF suites do not capture permanent geometric occupancy or
multi-terminal pipe groups, and we are not aware of a public 3D multi-group
pipe-routing suite with controlled difficulty and a constructive feasibility
certificate. We therefore build instances synthetically.

Each instance is a triple $(\mathcal{E},\{G_i\}_{i=1}^{m},\mathcal{W})$:
environment $\mathcal{E}$, $m$ groups with face terminals and uniform radius
$r$, and a conflict-free \textbf{witness} $\mathcal{W}$ proving generator-level
feasibility. Solvers never see $\mathcal{W}$ as a hint; it is used for
validation and as a \emph{reference length}, not an optimality bound.
Growing $\mathcal{W}$ first, then painting obstacles in its complement, yields
known-feasible clutter without coupling the suite to any solver under test.

\paragraph{Pipeline.}
(1)~Grow tree-like witness corridors with \textsc{AxisRay} or
\textsc{PlaneSlice} (groups attach in round-robin).
(2)~Paint static obstacles outside thickened witnesses with
\textsc{PillarField} or \textsc{CorridorRack}, never invading clearance $r$ on
$\mathcal{W}$.
(3)~Emit YAML (grid, obstacles, terminals, witness).
The product of two corridor methods and two obstacle families gives
\textbf{four datasets}, each with levels
\texttt{easy}/\texttt{middle}/\texttt{hard}
(Table~\ref{tab:difficulty}; $N{=}100$ successes per cell, domain $32^3$,
$r{=}0.5$).

\begin{table}[t]
  \centering
  \caption{Difficulty schedule ($m$ groups $\times$ $n_t$ terminals per group).
  Shared: domain $32^3$, $r{=}0.5$.}
  \label{tab:difficulty}
  \begin{tabular}{@{}lll@{}}
    \toprule
    Level & \textsc{AxisRay} ($m \times n_t$) & \textsc{PlaneSlice} ($m \times n_t$) \\
    \midrule
    easy   & $3\times 3$ & $3\times 3$ \\
    middle & $6\times 6$ & $8\times 8$ \\
    hard   & $9\times 9$ & $12\times 12$ \\
    \bottomrule
  \end{tabular}
\end{table}

\paragraph{Corridor generators.}
Both produce multi-terminal trees, not a single $s$--$t$ path; inter-group
corridors are obstacles for later growth.
\textsc{AxisRay} (preferred) grows an axis-aligned polyline tree by LOS scans
(no subdomain grid, no A$^\star$), with path/branch gaps and forced axis
changes at turns.
\textsc{PlaneSlice} grows on axis-aligned planes with \emph{planar} A$^\star$
only (no 3D fallback), using an $N{\times}N{\times}N$ subdomain layout and a
plane-gap within each group.
Pseudocode and clearance rules appear in Appendix~\ref{app:generators}.

\paragraph{Obstacle generators.}
Witness cells are thickened into a keep-out set $C$; obstacles never invade
$C$.
\textsc{PillarField} places large AABBs until target density
$\rho^\star{=}V_{\mathrm{obs}}/V_0$ (we use $\rho^\star{=}0.4$), sampling seeds
by distance to blocked cells.
\textsc{CorridorRack} dilates witnesses by $k$ Chebyshev steps
($k{=}2$ in reported runs) and meshes the complement as AABBs (density derived).
Details in Appendix~\ref{app:generators}.

\section{Evaluation Protocol}
\label{sec:eval}

Given $(\mathcal{E},\{G_i\})$ without $\mathcal{W}$, a solver must return
conflict-free paths for all topology tasks or fail within budget.
Reported runs fix topology to pure MST and low-level search to A$^\star$
(26-connected, Manhattan), and compare CBS / PBS / PP as in
Sec.~\ref{sec:method}.

\paragraph{Primary metric.}
Success rate under wall-clock budget $T$ (easy $30\,$s, middle $60\,$s, hard
$120\,$s):
\[
  \mathrm{SuccessRate}
  = \frac{1}{N}\bigl|\{ i : \text{instance $i$ solved within $T$}\}\bigr|.
\]
We also report failure reasons (\texttt{timeout} / \texttt{no\_solution}) and
success-vs-time curves. Fixed-budget success matches batch design use better
than optimality claims we do not make.

\paragraph{Secondary metrics.}
Runtime on successes; on successes,
$\mathrm{cost\_ratio}=\mathrm{cost}(\mathrm{sol})/\mathrm{cost}(\mathcal{W})$
as a descriptive quality signal (\emph{not} an optimality gap---$\mathcal{W}$
is constructive). Cost comparisons across algorithms use the common success
set to limit selection bias.

\paragraph{Reproducibility.}
The suite, batch runners, and JSONL summaries are in the public repository
\url{https://github.com/bladesaber/MAPF_Pipeline}
(branch \texttt{branch\_quan}, commit \texttt{cf0005685626dcaedb4c96a64fdca554312dfba1}).
Reported runs use \texttt{pipe\_design/include/benchmark/gen\_robust.sh},
\texttt{run\_bench.sh}, and \texttt{pipe\_design/scripts/analyze\_runs.py}.


\section{Experiments}
\label{sec:experiments}

We evaluate CBS, PBS, and PP on the four datasets of Sec.~\ref{sec:benchmark}
($N{=}100$ instances per cell). Shared solver settings match
Sec.~\ref{sec:eval}: pure MST topology, A$^\star$ / 26-connected / Manhattan,
wall-clock budgets $30$\,s / $60$\,s / $120$\,s for easy / middle / hard.
Tables report success rate under the level budget; hard Success@$T$ curves use
wall time with failures treated as $+\infty$.

\subsection{Success rate}
\label{sec:success}

Table~\ref{tab:success} summarises success rates.
On \textsc{AxisRay}, all three algorithms solve easy and middle almost
completely; on hard, CBS remains near-perfect ($99$--$100\%$) while PBS and PP
drop to roughly $77$--$81\%$ and $64$--$71\%$, respectively.
On \textsc{PlaneSlice}, easy remains saturated, but middle already separates
methods, and hard is strongly discriminating:
CBS reaches $56\%$ (\textsc{PillarField}) and $66\%$ (\textsc{CorridorRack}),
PBS only $7\%$ / $11\%$, and PP solves \emph{none} of the $200$ hard instances.
Obstacle family does not change the high-level ranking on
\textsc{PlaneSlice}; both painters support the same ordering
$\mathrm{CBS}\gg\mathrm{PBS}\gg\mathrm{PP}$.

\begin{table}[t]
  \centering
  \caption{Success rate (\%). Each cell: $N{=}100$. Budgets: easy $30$\,s,
  middle $60$\,s, hard $120$\,s.}
  \label{tab:success}
  \setlength{\tabcolsep}{4pt}
  \begin{tabular}{@{}llrrr@{}}
    \toprule
    Dataset & Level & CBS & PBS & PP \\
    \midrule
    \textsc{AxisRay}$\times$\textsc{PillarField}
      & easy   & 100 & 100 & 100 \\
      & middle & 100 &  99 &  99 \\
      & hard   &  99 &  81 &  71 \\
    \midrule
    \textsc{AxisRay}$\times$\textsc{CorridorRack}
      & easy   & 100 & 100 & 100 \\
      & middle & 100 &  98 &  98 \\
      & hard   & 100 &  77 &  64 \\
    \midrule
    \textsc{PlaneSlice}$\times$\textsc{PillarField}
      & easy   & 100 & 100 & 100 \\
      & middle &  98 &  90 &  75 \\
      & hard   &  56 &   7 &   0 \\
    \midrule
    \textsc{PlaneSlice}$\times$\textsc{CorridorRack}
      & easy   & 100 & 100 & 100 \\
      & middle & 100 &  96 &  80 \\
      & hard   &  66 &  11 &   0 \\
    \bottomrule
  \end{tabular}
\end{table}

\subsection{Runtime and Success@$T$}
\label{sec:runtime}

When successful, easy instances finish in tens of milliseconds (PP often
$\sim$1\,ms).
On \textsc{AxisRay} hard, CBS/PBS median success times stay below $2$\,s; PP
successes are much slower (median $\sim$38\,s).
On \textsc{PlaneSlice} hard, CBS median success time is on the order of
$46$--$61$\,s, and many failures are timeouts---the budget is binding.

Table~\ref{tab:success-at-t} reports hard Success@$T$ at $5$/$30$/$60$/$100$\,s
(budget in the runs is $120$\,s; the $100$\,s column is slightly conservative).
\textsc{AxisRay} hard is largely decided by $5$--$30$\,s for CBS/PBS.
\textsc{PlaneSlice} hard continues to accumulate CBS successes through $100$\,s
($50$--$62\%$), while PBS plateaus near $7$--$9\%$ and PP remains at $0\%$.
PBS/PP gains from $60$\,s to $100$\,s are small: structure and dead orders
dominate, not a missing few seconds.

\begin{table}[t]
  \centering
  \caption{Hard Success@$T$ (\%). Failures counted as $+\infty$ time.}
  \label{tab:success-at-t}
  \setlength{\tabcolsep}{3.5pt}
  \begin{tabular}{@{}llrrrr@{}}
    \toprule
    Dataset & Algo & $@5$\,s & $@30$\,s & $@60$\,s & $@100$\,s \\
    \midrule
    \textsc{AxisRay}$\times$\textsc{PillarField}
      & CBS & 96 & 99 & 99 & 99 \\
      & PBS & 78 & 81 & 81 & 81 \\
      & PP  & 16 & 31 & 52 & 67 \\
    \midrule
    \textsc{AxisRay}$\times$\textsc{CorridorRack}
      & CBS & 94 & 100 & 100 & 100 \\
      & PBS & 76 &  76 &  77 &  77 \\
      & PP  & 11 &  27 &  49 &  61 \\
    \midrule
    \textsc{PlaneSlice}$\times$\textsc{PillarField}
      & CBS &  0 &  8 & 28 & 50 \\
      & PBS &  0 &  2 &  6 &  7 \\
      & PP  &  0 &  0 &  0 &  0 \\
    \midrule
    \textsc{PlaneSlice}$\times$\textsc{CorridorRack}
      & CBS &  0 & 17 & 43 & 62 \\
      & PBS &  0 &  1 &  5 &  9 \\
      & PP  &  0 &  0 &  0 &  0 \\
    \bottomrule
  \end{tabular}
\end{table}

\subsection{Failures and path cost}
\label{sec:fail-cost}

On \textsc{PlaneSlice} hard, CBS failures are almost all timeouts
($44$ and $34$ of $100$).
PBS mixes timeouts with a few \texttt{no\_solution} outcomes; PP fails with a
large share of both timeout and exhausted order neighbourhoods
($\sim$33--34 \texttt{no\_solution} per obstacle family).
On \textsc{AxisRay} hard, PBS/PP failures are mostly \texttt{no\_solution}
rather than timeout, consistent with brittle orders once corridors interact.

Median $\mathrm{cost\_ratio}$ on successes is typically below $1$
(often $\sim$0.53--0.76 on hard non-trivial cells): constructive witnesses are
feasible but loose.
We treat $\mathrm{cost\_ratio}$ as a descriptive length signal only, not as an
optimality gap.


\section{Related Work}
\label{sec:related}

\textbf{Multi-agent pathfinding.}
Grid MAPF uses high-level controllers such as CBS~\cite{sharon2012cbs},
PBS~\cite{ma2019pbs}, and priority planning~\cite{felner2017socs}.
Agents occupy cells temporarily; conflicts are vertex- or edge-based at explicit
timesteps.
Stern et al.~\cite{stern2019mapf} unify MAPF terminology and publish grid
benchmarks for spatiotemporal planning.
We reuse the same \emph{control patterns}---constraint trees, priority DAGs,
total orders---but under permanent geometric occupancy, dual-witness clearance
conflicts, and multi-terminal groups.
Our suite targets design-time 3D layout rather than concurrent motion on
MovingAI-style MAPF maps~\cite{stern2019mapf}.

\textbf{3D pipe routing and MAPF transfer.}
Belov et al.~\cite{belov2020pr} draw the analogy between MAPF blocked cells in
$x$--$y$--$t$ and pipe routing in $x$--$y$--$z$, and adapt CBS and PBS to
industrial PR instances with one start and one goal per pipe.
They evaluate on plant-scale and smaller synthetic cases, but do not provide a
public multi-group benchmark with constructive feasibility witnesses or a
systematic comparison including priority planning.
We extend the MAPF-to-routing line to \emph{grouped} multi-terminal topology,
explicit permanent occupancy, and a reproducible four-dataset suite
(Sec.~\ref{sec:benchmark}).

\textbf{Automatic pipe routing.}
Blokland et al.~\cite{blokland2023pipe} survey automatic pipe routing for
large architectures (ships, chemical plants), where sequential CAD-style routing
and mixed-integer or sampling methods dominate and full automation remains
difficult.
That literature rarely expresses cross-pipe interaction as MAPF-style branching
and seldom releases controlled 3D instance generators with certificates.
Our contribution is closer to a MAPF benchmark paper~\cite{stern2019mapf} than
to a single-application optimiser: fixed protocol, open scripts, and baseline
success rates on $N{=}100$ instances per cell.

\textbf{Application context.}
Industrial PR motivates the abstraction~\cite{belov2020pr,blokland2023pipe};
our intro uses AM-oriented multi-group valve layout as a representative case.
Reported experiments remain on constructive $32^3$ grids so that difficulty and
metrics are comparable across CBS, PBS, and PP.


\section{Discussion}
\label{sec:discussion}

The benchmark is not uniformly hard.
\textsc{AxisRay} easy and middle saturate for all three controllers; only on
hard do PBS and PP fall behind CBS, and even then CBS stays near $100\%$.
\textsc{PlaneSlice}---especially hard with many groups---is where methods
separate: CBS reaches $56$--$66\%$ success while PBS drops to single digits
and PP solves none of the $200$ hard instances under the same budget.
That pattern matches the problem structure: once prior pipes are permanent
obstacles, order- and priority-based search can lock in early commitments,
whereas CBS can branch on geometric conflicts and add localized negative
constraints.
Obstacle family (\textsc{PillarField} vs.\ \textsc{CorridorRack}) does not
change the ranking on \textsc{PlaneSlice}, which suggests the stress test
comes from corridor layout and group count rather than a particular clutter
model.

Several scope limits follow directly.
Instances are synthetic $32^3$ grids with uniform radius; we do not evaluate on
B-rep valve bodies, multi-radius mixtures, or manufacturing criteria such as
bend limits or supportability.
Constructive witnesses certify generator feasibility only---$\mathrm{cost\_ratio}$
below $1$ must not be read as an optimality gap.
We report CBS, PBS, and PP as baselines under one fixed protocol, not as an
exhaustive comparison to commercial CAD routers or continuous shape
optimisers; post-layout smoothing remains orthogonal follow-on work.

\section{Conclusion}
\label{sec:conclusion}

We formalized multi-group 3D pipe routing under permanent geometric occupancy,
introduced a constructive benchmark with four datasets and controlled difficulty,
and measured fixed-budget success for CBS, PBS, and priority planning on
$N{=}100$ instances per cell.
The suite and reported runs are intended as a reproducible starting point for
stronger geometric multi-group planners and for coupling discrete layout search
to manufacturing-aware refinement.

\appendix

\section{Generator Details}
\label{app:generators}

This appendix records constructive generators used in Sec.~\ref{sec:benchmark}.
It is intended for reproducibility; the main text only needs the high-level
behaviour.

\subsection{\textsc{AxisRay}}
\label{app:axis-ray}

\textsc{AxisRay} uses LOS scans with rollback on failed seed/attach attempts.
Hard rules include: no entry into written path cells; Chebyshev gap
\texttt{path\_gap} except along the spine of the edge being left (requires
\texttt{branch\_gap}$>$\texttt{path\_gap}); consecutive segments change axis;
branch points stay farther than \texttt{branch\_gap} from existing waypoints.

\begin{algorithm}[t]
\caption{\textsc{AxisRay}: generate witness corridors}
\label{alg:axis-ray}
\begin{algorithmic}[1]
\Require domain size, radius $r$, groups $\{n_t^{(g)}\}$, waypoint budgets, gaps
\Ensure per-group corridors or \textbf{fail}
\State $\mathrm{Occ}\gets\emptyset$; $\mathrm{WP}[g]\gets\emptyset$ for all groups $g$
\While{some group needs more face terminals}
  \For{group $g$ in round-robin order}
    \If{$g$ has no corridor}
      \State sample face terminal $s$; grow seed chain with
             $\texttt{init\_waypoints\_num}{-}2$ interior LOS segments
      \State final LOS segment to a domain face; record waypoints
      \State \textbf{on fail:} rollback $\mathrm{Occ}$ / paths for this attempt
    \Else
      \State pick branch point $p$ on $g$'s tree with
             $\mathrm{dist}(p,\mathrm{WP}[g])>\texttt{branch\_gap}$
      \State pick side-exit direction (not along the current edge axis)
      \State grow attach chain ($\texttt{attach\_waypoints\_num}$) to a new face
      \State \textbf{on fail:} rollback this attach
    \EndIf
  \EndFor
  \If{no progress after max attempts}
    \State \Return \textbf{fail}
  \EndIf
\EndWhile
\State \Return corridors
\end{algorithmic}
\end{algorithm}

\subsection{\textsc{PlaneSlice}}
\label{app:plane-slice}

Every search is restricted to the active axis plane (2D); there is no fallback
to full 3D A$^\star$. Same-group corridors off the plane are ignored during
search; other groups on the plane are obstacles. Same-axis planes within a
group obey a plane-gap.

\begin{algorithm}[t]
\caption{\textsc{PlaneSlice}: generate witness corridors}
\label{alg:plane-slice}
\begin{algorithmic}[1]
\Require domain, $r$, subdomain $N$, plane gap, targets $\{n_t^{(g)}\}$
\Ensure per-group corridors or \textbf{fail}
\State build subdomain partition; $\mathrm{Planes}[g]\gets\emptyset$
\While{some group needs more terminals}
  \For{group $g$ in round-robin order}
    \If{$g$ empty}
      \State sample valid axis-plane $P$; place two face terminals $s,t$ on $P$
      \State planar A$^\star$ on $P$ only; record path and $P$
    \Else
      \State sample plane type ${\neq}$ last plane; derive $P$ from a cell of $g$
      \State multi-source starts $\gets$ all cells of $g$ lying on $P$
      \State pick new face goal on $P$; planar multi-source A$^\star$ on $P$
      \State on success: append path, record $P$
    \EndIf
  \EndFor
\EndWhile
\State \Return corridors
\end{algorithmic}
\end{algorithm}

\subsection{\textsc{PillarField} and \textsc{CorridorRack}}
\label{app:obstacles}

Let $C$ be witness cells thickened by \texttt{corridor\_margin} (and $r$).
Eligible volume $V_0$ is the complement of $C$ at paint start.

\begin{algorithm}[t]
\caption{\textsc{PillarField}: paint AABB obstacles}
\label{alg:pillar}
\begin{algorithmic}[1]
\Require instance with witness, $\rho^\star$, edge bounds, \texttt{dist\_cap}
\State mark thickened witness as blocked; build Chebyshev distance field $d(\cdot)$
\State $V_0\gets$ count of free cells; $V_{\mathrm{obs}}\gets 0$
\While{$V_{\mathrm{obs}}/V_0 < \rho^\star$ and progress}
  \State sample seed $c$ with weight $\propto \min(d(c),\texttt{dist\_cap})$
  \State expand AABB from $c$ (greedy / isometric); reject if too thin
  \State among candidates, commit the largest-volume box into blocked
  \State update $d(\cdot)$ from newly blocked cells; $V_{\mathrm{obs}}\mathrel{+}=$ volume
\EndWhile
\State emit committed boxes as static obstacles
\end{algorithmic}
\end{algorithm}

\begin{algorithm}[t]
\caption{\textsc{CorridorRack}: complement of dilated witnesses}
\label{alg:rack}
\begin{algorithmic}[1]
\Require instance with witness, dilate steps $k$
\State $F \gets$ Chebyshev dilate of all witness cells by $k$
\State $\mathrm{ObsCells} \gets$ domain $\setminus F$
\State merge $\mathrm{ObsCells}$ into large AABBs; append to static obstacles
\State record derived $\rho \gets V_{\mathrm{obs}} / V_{\mathrm{domain}}$
\end{algorithmic}
\end{algorithm}

\end{document}